%% file: emnlp2018_main.tex
\documentclass[11pt,a4paper]{article}
\usepackage{authblk}
\usepackage[hyperref]{emnlp2018}
\usepackage{times}
\usepackage{latexsym}

\usepackage[T2A, T1]{fontenc}
\usepackage[utf8]{inputenc}

\usepackage{url}
\usepackage{graphicx}
\usepackage{multirow}
\usepackage{amsthm}
\usepackage{amssymb}
\usepackage{amsmath}
\usepackage{amsfonts}
\usepackage{color}
\usepackage{mdwlist}
\usepackage{microtype}
\usepackage{tabularx}
\usepackage{booktabs}
\usepackage{subcaption}

\usepackage{color,colortbl}

\definecolor{Gray}{gray}{0.9}
\usepackage{enumitem,amssymb}
\newlist{todolist}{itemize}{2}
\setlist[todolist]{label=$\square$}
\usepackage{pifont}
\usepackage[russian,english]{babel}

\aclfinalcopy 
\def\aclpaperid{1036} 

\title{Adversarial Propagation and Zero-Shot Cross-Lingual Transfer \\ of Word Vector Specialization}

\author{\bf Edoardo M. Ponti$^1$\thanks{{ } Both authors equally contributed to this work.}, Ivan Vuli\'{c}$^{1*}$, Goran Glavaš$^2$, {\bf Nikola Mrkšić$^3$, Anna Korhonen}}
\affil[1]{Language Technology Lab, DTAL, University of Cambridge}
\affil[2]{Data and Web Science Group, University of Mannheim} 
\affil[3]{PolyAI}
\affil[]{\texttt {\{ep490,iv250,alk23\}@cam.ac.uk}}
\affil[2]{\texttt {goran@informatik.uni-mannheim.de}}
\affil[3]{\texttt {nikola@poly-ai.com}}

\date{}

\begin{document}
\maketitle
\begin{abstract}
Semantic \textit{specialization} is a process of fine-tuning pre-trained distributional word vectors using external lexical knowledge (e.g., WordNet) to accentuate a particular semantic relation in the specialized vector space. While post-processing specialization methods are applicable to arbitrary distributional vectors, they are limited to updating only the vectors of words occurring in external lexicons (i.e., \textit{seen words}), leaving the vectors of all other words unchanged. We propose a novel approach to specializing the full distributional vocabulary. Our \textit{adversarial post-specialization} method propagates the external lexical knowledge to the full distributional space. We exploit words seen in the resources as training examples for learning a global specialization function. This function is learned by combining a standard $L_2$-distance loss with a adversarial loss: the adversarial component produces more realistic output vectors. We show the effectiveness and robustness of the proposed method across three languages and on three tasks: word similarity, dialog state tracking, and lexical simplification. We report consistent improvements over distributional word vectors and vectors specialized by other state-of-the-art specialization frameworks. Finally, we also propose a cross-lingual transfer method for \textit{zero-shot specialization} which successfully specializes a full target distributional space without any lexical knowledge in the target language and without any bilingual data.
\end{abstract}


\section{Introduction}
\label{s:intro}
\input{01_intro}

\section{Methodology}
\label{s:methodology}
\input{03_methodology}

\section{Experimental Setup}
\label{s:experimental}
\input{04_experimental}

\section{Results and Discussion}
\label{s:results}
\input{05_results}

\section{Related Work}
\label{s:related}
\input{02_related}

\section{Conclusion and Future Work}
\label{06_conclusions}
\input{06_conclusions }

\section*{Acknowledgements}
This work is supported by the ERC Consolidator Grant LEXICAL (no 648909). 

\bibliographystyle{acl_natbib}
\bibliography{emnlp2018_refs}
\end{document}

%% file: 01_intro.tex
Word representation learning is a mainstay of modern Natural Language Processing (NLP), and its usefulness has been proven across a wide spectrum of NLP applications \cite[\textit{inter alia}]{Collobert:2011jmlr,Chen:2014emnlp,Melamud:2016naacl}. Standard \textit{distributional} word vector models are grounded in the distributional hypothesis \cite{Harris:1954}, that is, they leverage information about word co-occurrences in large text corpora \cite{Mikolov:2013nips,Pennington:2014emnlp,Levy:2014acl,Bojanowski:2017tacl}. This dependence on contextual signal results in a well-known tendency to conflate semantic similarity with other types of semantic association \cite{Hill:2015cl,Schwartz:2015conll,Vulic:2017conll} in the induced word vector spaces.\footnote{For instance, it is difficult to discern synonyms from antonyms in distributional vector spaces: this has a negative impact on language understanding tasks such as statistical dialog modeling or text simplification \cite{Glavas:2015acl,Faruqui:2015naacl,Mrksic:2016naacl,Kim:2016slt}}


A common remedy is to move beyond purely unsupervised word representation learning, in a process referred to as \textit{semantic specialization} or \textit{retrofitting}. Specialization methods exploit lexical knowledge from external resources, such as WordNet \cite{Fellbaum:1998wn} or the Paraphrase Database \cite{Ganitkevitch:2013naacl} to refine the semantic properties of pre-trained vectors and \textit{specialize} the distributional spaces for a particular relation, e.g., synonymy (i.e., true similarity) \cite{Faruqui:2015naacl,Mrksic:2017tacl} or hypernymy \cite{Nickel:2017arxiv,Nguyen:2017emnlp,Vulic:2018lear}.

\begin{figure}[!t]
\begin{center}
\includegraphics[width=0.88\linewidth]{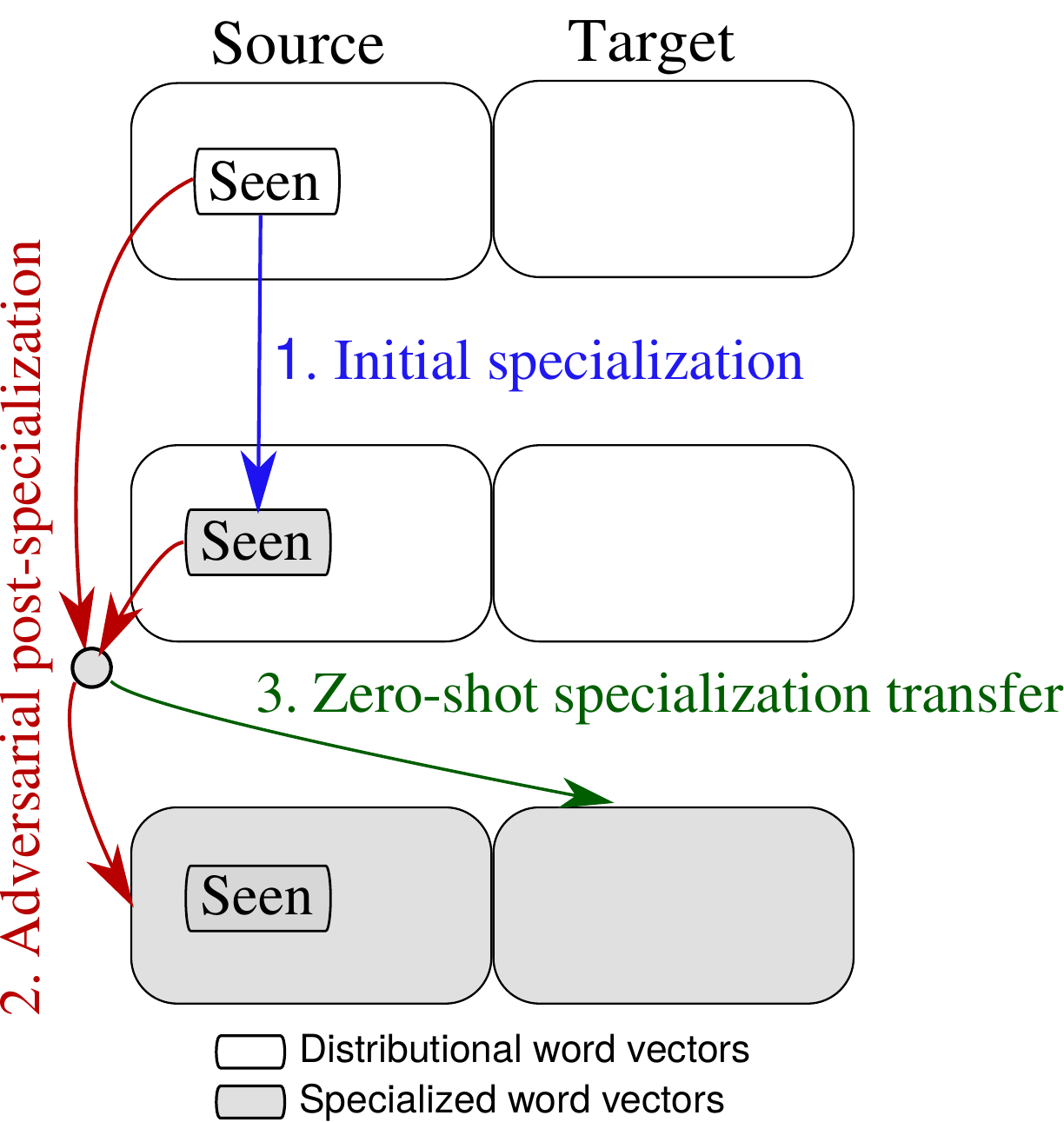}
\end{center}
\caption{High-level illustration of the adversarial post-specialization process and cross-lingual zero-shot specialization, described in detail in \S\ref{s:methodology}.}
\label{fig:illustration}
\end{figure}

The best-performing specialization models (cf. \citealt{Mrksic:2017tacl}) are deployed as \textit{post-processors of the vector space}: distributional vectors are \textit{fine-tuned} to satisfy linguistic constraints extracted from external resources to offer improved support to downstream NLP applications \cite{Faruqui:2016thesis}. Such models are versatile as they can be applied to arbitrary distributional spaces, but they have a major drawback: they \textit{locally} update only vectors of words present in linguistic constraints (i.e., \textit{seen} words), whereas vectors of all other (i.e., \textit{unseen}) words remain intact (see Figure~\ref{fig:illustration}).

\newcite{Vulic:2018naaclpost} have recently proposed a model which, based on the updates of vectors of seen words, learns a global specialization function that can be applied to the large subspace of unseen words. Their global method, termed \textit{post-specialization} and implemented as a deep feed-forward network, effectively specializes \textit{all} distributional vectors. 

In this paper, we propose a new approach to post-specialization which addresses the following two research questions: \textbf{a)} Is it possible to use a more sophisticated learning approach to yield more realistic specialized vectors for the full vocabulary? \textbf{b)} Given that specialization methods inherently require a large number of constraints, is it possible to specialize distributional word vectors where such resources are scarce or non-existent? Our novel model learns the global specialization function by casting the feed-forward specialization network as a generator component of an adversarial architecture, see Figure~\ref{fig:auxgan}. The corresponding discriminator component learns to discern original specialized vectors (produced by any \textit{local} specialization model) from vectors produced by transforming distributional vectors with the feed-forward post-specialization network (i.e., the generator). 

We show that the proposed adversarial model yields state-of-the-art performance on standard word similarity benchmarks, outperforming the  post-specialization model of \newcite{Vulic:2018naaclpost}. We further demonstrate the effectiveness of the proposed model in two downstream tasks: lexical text simplification and dialog state tracking. Finally, we demonstrate that, by coupling our adversarial specialization model with any unsupervised model for inducing bilingual vector spaces, such as the algorithm proposed by \newcite{Conneau:2018iclr}, we can successfully perform zero-shot language transfer of the specialization, that is, we can specialize distributional spaces of languages without any linguistic constraints in those languages, and without any bilingual data.    


%% file: 03_methodology.tex
The {post-specialization} procedure \cite{Vulic:2018naaclpost} is a two-step process. First, a subspace of vectors for words observed in external resources is fine-tuned using any off-the-shelf specialization model, such as the original retrofitting model \cite{Faruqui:2015naacl}, counter-fitting \cite{Mrksic:2016naacl}, dLCE \cite{Nguyen:2016acl}, or state-of-the-art \textsc{attract-repel (ar)} specialization \cite{Mrksic:2017tacl,Vulic:2017acl}. We outline the initial specialization algorithms in \S\ref{ssec:ar}. In the second step, the initial specialization is propagated to the entire vocabulary, including words not observed in the resources, relying on an adversarial architecture augmented with a distance loss. This adversarial post-specialization model, compatible with any specialization model, is described in \S\ref{ssec:gan}. 

Finally, in \S\ref{ssec:biling}, we introduce a \textit{cross-lingual zero-shot specialization model} which transfers the specialization to a target language without any lexical resources.  An overview of the proposed methodology from this section is provided in Figure~\ref{fig:illustration}. 


\subsection{Initial Specialization}
\label{ssec:ar}
\paragraph{Linguistic Constraints.} 
Adopting the nomenclature from \newcite{Mrksic:2017tacl}, post-processing models are generally guided by two broad sets of constraints: \textbf{1)} \textsc{attract} constraints specify which words should be close to each other in the fine-tuned vector space (e.g.\ synonyms like \textit{graceful} and \textit{amiable}); \textbf{2)} \textsc{repel} constraints describe which words should be pulled away from each other (e.g.\ antonyms like \textit{innocent} and \textit{sinful}). Earlier post-processors \cite{Faruqui:2015naacl,Jauhar:2015,Wieting:2015tacl} operate only with \textsc{attract} constraints, and are thus not suited to model both aspects contributing to the specialization process. 

We first outline the state-of-the-art \textsc{attract-repel} specialization model \cite{Mrksic:2017tacl} which leverages both sets of constraints. Here, we again stress two important aspects relevant to our post-specialization model: \textbf{a)} all initial specialization models fine-tune only representations for the subspace of words seen in the external constraints, while all other words remain unaffected by specialization; \textbf{b)} post-specialization is not tied to \textsc{attract-repel} in particular; it is applicable on top of any other post-processor.\footnote{We have empirically validated the robustness of the proposed adversarial post-specialization by applying it also on top of other post-processing methods: retrofitting \cite{Faruqui:2015naacl} and counter-fitting \cite{Mrksic:2016naacl}. For brevity, we only report the (best) results with \textsc{attract-repel}, the best-performing initial/local specialization model.} 

\paragraph{Specialization of Seen Words.}
The key idea is to inject the knowledge from linguistic constraints into pre-trained distributional word vectors. Given a set $A$ of \textsc{attract} word pairs and a set $R$ of \textsc{repel} word pairs, each word pair $(v_l, v_r)$ from the vocabulary $\mathcal{V}_s$ of seen words present in these sets can be represented as a vector pair ($\textbf{x}_l, \textbf{x}_r)$.

The optimization is driven by mini-batches of \textsc{attract} pairs $\mathcal{B}_A$ (batch size $k_A$), and of \textsc{repel} pairs $\mathcal{B}_R$ (size $k_R$). For both of these, two sets of negative example pairs of equal size are drawn from the $2(k_A + k_R)$ vectors occurring in $\mathcal{B}_A$ and $\mathcal{B}_R$. This defines the mini-batches $T_A ({B}_A) = [(\mathbf{t}_l^1, \mathbf{t}_r^1) \dots (\mathbf{t}_l^{k_A}, \mathbf{t}_r^{k_A})]$ and $T_R ({B}_R) = [(\mathbf{t}_l^1, \mathbf{t}_r^1) \dots (\mathbf{t}_l^{k_R}, \mathbf{t}_r^{k_R})]$. Negative examples $\textbf{t}_l$ and $\textbf{t}_r$ for \textsc{attract} (or \textsc{repel}) pairs are the nearest (or farthest) neighbours by cosine similarity to $\textbf{x}_l$ and $\textbf{x}_r$, respectively. They ensure that the paired vectors for words in the constraints are closer to each other (or more distant for antonyms) than to their respective negative examples. 

The overall objective function consists of three terms. The first term pulls \textsc{attract} pairs together:

\vspace{-2.5mm}
{\footnotesize
\begin{multline}
Att(\mathcal{B}_A, \mathcal{T}_A) = \sum_{i=1}^{k_A} \Big[ \tau \big( \delta_A + \mathbf{x}_l^i \mathbf{t}_l^i - \mathbf{x}_l^i \mathbf{x}_r^i \big) + \\[-0.1mm]
+ \tau \big( \delta_A + \mathbf{x}_r^i \mathbf{t}_r^i - \mathbf{x}_l^i \mathbf{x}_r^i \big) \Big]
\end{multline}}%

\noindent $\tau(z) = \text{max}(0, z)$ is the standard rectifier \citep{Nair:2010icml}. $\delta_A$ is the \textsc{attract} margin: it specifies the tolerance for the difference between the two distances (with the other pair member and with the negative example). The second term, $Rep(\mathcal{B}_R, \mathcal{T}_R)$, is similar but now pushes \textsc{repel} pairs away from each other, relying on the \textsc{repel} margin $\delta_R$:

{\footnotesize
\begin{multline}
Rep(\mathcal{B}_R, \mathcal{T}_R) = \sum_{i=1}^{k_R} [ \tau \left( \delta_R - \mathbf{x}_l^i \mathbf{t}_l^i + \mathbf{x}_l^i \mathbf{x}_r^i \right) + \\
+ \tau \left( \delta_R - \mathbf{x}_r^i \mathbf{t}_r^i + \mathbf{x}_l^i \mathbf{x}_r^i \right) ]
\end{multline}}%

\noindent
The final term is tasked to \textit{preserve} the quality of the original vectors through $L_2$-regularization:

\vspace{-0.5mm}
{\footnotesize
\begin{align}
Pre(\mathcal{B}_A, \mathcal{B}_R) = \sum_{\textbf{x}_i \in \mathcal{B}_A \cup \mathcal{B}_R} \lambda_P || \textbf{y}_i - \textbf{x}_i ||_2
\end{align}}%

\noindent $\textbf{y}_i$ is the vector specialized from the original distributional vector $\textbf{x}_i$, and $\lambda_P$ is a regularization hyper-parameter. The optimizer finally minimizes the following objective: $ \mathcal{L}_{AR} = Att(\mathcal{B}_A, \mathcal{T}_A) + Rep(\mathcal{B}_R, \mathcal{T}_R) + Pre(\mathcal{B}_A, \mathcal{B}_R)$.

\begin{figure*}[!t]
\centering
\includegraphics[width=0.85\textwidth, trim={0 3.3cm 0 3.2cm},clip]{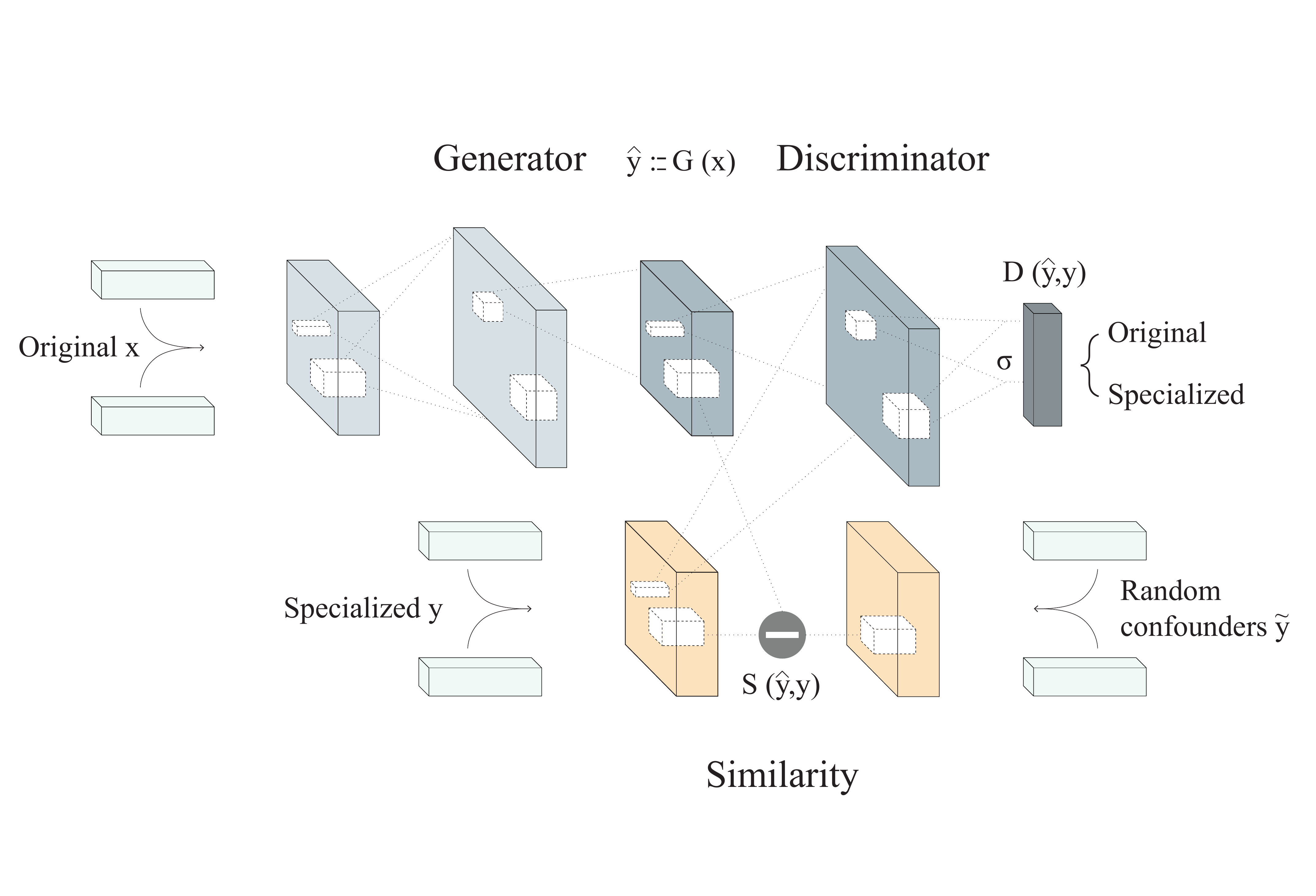}
\caption{Architecture of the AuxGAN: an adversarial generator-discriminator loss (above) is combined with a max-margin $L_2$-distance loss with random confounders (below).}
\label{fig:auxgan}
\end{figure*}

\subsection{Adversarial Post-Specialization}
\label{ssec:gan}

\paragraph{Motivation.} 
The AR method affects only a subset of the full vocabulary $\mathcal{V}$, and consequently only a (small) subspace of the original space $\textbf{X}$ (see Figure \ref{fig:illustration}). In particular, it specializes the embeddings $\textbf{X}_s$ corresponding to $\mathcal{V}_s$, the vocabulary of words observed in the constraints. It leaves the embeddings $\textbf{X}_u$ corresponding to all other (unseen) words $\mathcal{V}_u$ identical.

Nevertheless, the perturbation underwent by the original observed embeddings can provide evidence about the general effects of specialization. In particular, it allows to learn a global mapping function $f: \textbf{X} \in \mathbb{R}^d \rightarrow \textbf{Y} \in \mathbb{R}^d$ for \textit{d}-dimensional vectors. The parameters for this function can be trained in a supervised fashion from pairs of original and initially specialized word embeddings ($\textbf{x}_i^{(s)}$, $\textbf{y}_i^{(s)}$) from $\mathcal{V}_s$, as illustrated by Figure~\ref{fig:auxgan}. Subsequently, the mapping can be applied to distributional word vectors $\textbf{x}_u$ from the vocabulary of unseen words $\mathcal{V}_u$ to predict $\textbf{\^y}_u$, their specialized counterpart. This procedure, called post-specialization, effectively propagates the information stored in the external constraints to the entire word vector space.

However, this mapping should not just model the inherent transformation, but also ensure that the resulting vector is `natural'. In particular, assuming that word representations lie on a manifold, the mapping should return one of its values. The intuition behind our formulation of the training objective is that: \textbf{a)} an $L_2$-distance loss can retrieve a faithful mapping whereas \textbf{b)} an adversarial loss can prevent unrealistic outputs, as already proven in the the visual domain \cite{pathak2016context,ledig2017photo,Odena:2017icml}.

\paragraph{Objective Function.}
The pairs of original and specialized embeddings for seen words allow to train the global mapping function. In principle, this can be any differentiable parametrized function $G(\textbf{x};\theta_G)$. \citet{Vulic:2018naaclpost} showed that non-linear functions ensure a better mapping than linear transformations which seem inadequate to mimic the complex perturbations of the specialization process, guided by possibly millions of pairwise constraints. Our preliminary experiments corroborate this intuition. 
Thus, in this work we also opt for implementing $G(\textbf{x};\theta_G)$ as a deep neural network. Each of the \textit{l} hidden layers of size \textit{h} non-linearly transforms its input. The output layer is a linear transformation into the prediction $\textbf{\^y} \in \mathcal{R}^{d}$.

The parameters $\theta_G$ are learned by minimizing the $L_2$ distance between the training pairs. In particular, the loss is a contrastive margin-based ranking loss with negative sampling (MM) as proposed by \citet[\textit{inter alia}]{Weston:2011ijcai}. The gist of this loss is that the first component increases the cosine similarity \textit{cos} of predicted and initially specialized vectors of the same word up to a margin $\delta_{MM}$. On the other hand, the second component encourages the predicted vectors to distance themselves from $k$ \textit{random confounders}. These are negative examples sampled uniformly from the batch $\mathcal{B}$ excluding the current vector:

\vspace{-3mm}
{\footnotesize
\begin{multline} \label{eq:mm}
\mathcal{L}_{MM} = \sum_{i=1}^{||\mathcal{V}_s||} \sum_{j=1 | j \neq i}^{k} \tau [\delta_{MM} - \cos(G(\mathbf{x}_i^{(s)};\theta_G), \mathbf{y}_i^{(s)}) + \\[-0.5mm]
+ \cos(G(\mathbf{x}_i^{(s)};\theta_G), \mathbf{y}_j^{(s)})]
\end{multline}}%

\noindent One of the original contributions of this work is combining the $L_2$ distance with an adversarial loss, resulting in an auxiliary-loss Generative Adversarial Network (AuxGAN) as shown in Figure~\ref{fig:auxgan}. The role of the adversarial component, as mentioned above, is to `soften' the mapping and guarantee realistic outputs from the target distribution.

The mapping can be considered a generator $G(\textbf{x}|\theta_G)$. On top of this, a discriminator $D(\textbf{x} |\theta_D)$, implemented also as a multi-layer neural net, tries to distinguish whether a vector is sampled from the predicted vectors or the \textsc{ar}-specialized vectors. Its output layer performs binary classification through \textit{softmax}. The objective minimizes the loss $\mathcal{L}_D$: 

\vspace{-3mm}
{\footnotesize
\begin{multline} \label{eq:ld}
\mathcal{L}_{D} = - \sum_{i=1}^{n} \log P (\text{specialized} = 0 | G(\mathbf{x}_i; \theta_G); \theta_D ) - \\[-0.1mm] -  \sum_{i=1}^{m} \log P (\text{specialized} = 1 | \mathbf{y}_i; \theta_D )
\end{multline}}%

\noindent In a two-player game \cite{goodfellow2014generative}, the generator is trained to fool the discriminator by maximizing $ \text{log} ( 1 - P (0 | G(\mathbf{x}_i; \theta_G); \theta_D ))$. However, to avoid vanishing gradients of $G$ early on, the loss $\mathcal{L}_G$ is reformulated by swapping the labels of Eq.~\eqref{eq:ld} as follows:

\vspace{-2.5mm}
{\footnotesize
\begin{multline} \label{eq:lg}
\mathcal{L}_{G} = - \sum_{i=1}^{n} \log P (\text{specialized} = 1 | G(\mathbf{x}_i; \theta_G); \theta_D ) - \\[-0.1mm] -  \sum_{i=1}^{m} \log P (\text{specialized} = 0 | \mathbf{y}_i; \theta_D )
\end{multline}}%

\noindent During the optimization procedure through stochastic gradient descent, we alternate among \textit{s} steps for $\mathcal{L}_{D}$, one step for $\mathcal{L}_{G}$, and one step for $\mathcal{L}_{\mathit{MM}}$ to avoid the overfitting of $D$. 
The reason why $s \geq 1$ is that $D$ can be kept close to a minimum of its loss function by updating $G$ less frequently. 

\subsection{Zero-shot Transfer to Other Languages}
\label{ssec:biling}
Once the AuxGAN has learned a global mapping function $G(\textbf{x}; \theta_G)$ in a resource-rich language, it can be directly applied to unseen words. In this work, we propose a method to additionally post-specialize the whole vocabulary $\mathcal{V}_t$ of a resource-poor target language. We assume a real-world scenario where no target language constraints are available to specialize it directly. 

What is more, we assume that no bilingual data or dictionaries are available either. Hence, we rely on unsupervised cross-lingual word embedding induction, and in particular on \newcite{Conneau:2018iclr}'s method. By virtue of these assumptions, there is no limitation to the range of potential target languages that can be specialized. Incidentally, please note that the proposed transfer method is equally applicable on top of other cross-lingual word embedding induction methods. These may require more bilingual supervision to learn the cross-lingual vector space.\footnote{See the recent survey papers on cross-lingual word embeddings and their typology \cite{Upadhyay:2016acl,Vulic:2016acl,Ruder:2017arxiv}}


After learning the shared cross-lingual word embedding space in an unsupervised fashion \cite{Conneau:2018iclr}, the global post-specialization function learnt on the seen source language vectors is applied to the target language vectors, since they lie in the same shared space (see Figure \ref{fig:illustration} again). 
By virtue of the transfer, linguistic constraints in the source language can enhance the distributional vectors of target language vocabularies.

\newcite{Conneau:2018iclr} learn a shared cross-lingual vector space as follows. They first learn a coarse initial mapping between two monolingual embedding spaces in two different languages through a GAN where the generator is a linear transformation with an orthogonal matrix $\textbf{\^W}$. Its loss is identical to Eq.~\eqref{eq:ld} and Eq.~\eqref{eq:lg}, but unlike our AuxGAN model it discriminates between embeddings drawn from the source language 
and the target language distributions. Using the shared space, they extract for each source vector the closest target vector according to a distance metric designed to mitigate the hubness problem \cite{Radovanovic:2010jmlr}, the Cross-Domain Similarity Local Scaling (CSLS). 

This creates a bilingual synthetic dictionary that allows to further refine the coarse initial mapping. In particular, the optimal parameters for the linear mapping minimizing the $L_2$-distance between source-target pairs are provided by the closed-form Procrustes solution \citep{schonemann1966generalized} based on singular value decomposition (SVD):

{\footnotesize
\begin{multline}
\textbf{\^W} = \text{arg min}_{W} ||\textbf{W} \, \mathbf{X}_{t} - \mathbf{X}_{s}||_F = \textbf{UV}^\top \\ \mathbf{U\Sigma V}^\top = \text{SVD}(\mathbf{X}_{t} \mathbf{X}_{s}^\top)
\end{multline}}%

\noindent
where $|| \cdot ||_F$ is the Frobenius norm. After mapping the original target embeddings into the shared space with this method, we post-specialize them with the function outlined in \S\ref{ssec:gan}, learnt on the source language. This yields the specialized target vectors $\textbf{\^Y}_t = G(\textbf{\^W} \, \mathbf{X}_t ; \theta_G)$.


%% file: 04_experimental.tex
\paragraph{Distributional Vectors.}
We estimate the robustness of adversarial post-specialization by experimenting with three widely used collections of distributional English vectors. 1) \textsc{sgns-w2} vectors are trained on the cleaned and tokenized Polyglot Wikipedia \cite{AlRfou:2013conll} using Skip-Gram with Negative Sampling (SGNS) \cite{Mikolov:2013nips} by \newcite{Levy:2014acl} with bag-of-words contexts (window size is 2). 2) \textsc{glove-cc} are GloVe vectors trained on the Common Crawl \cite{Pennington:2014emnlp}. 3) \textsc{fasttext} are vectors trained on Wikipedia with a SGNS variant that builds word vectors by summing the vectors of their constituent character n-grams \cite{Bojanowski:2017tacl}. All vectors are $300$-dimensional.\footnote{Experiments with other standard word vectors, such as \textsc{context2vec} \cite{Melamud:2016conll} and dependency-based embeddings \cite{Bansal:2014acl} show similar trends and lead to same conclusions.}


\paragraph{Constraints and Initial Specialization.}
We experiment with the sets of linguistic constraints used in prior work \cite{Zhang:2014emnlp,Ono:2015naacl,Vulic:2018naaclpost}. These constraints, extracted from WordNet \cite{Fellbaum:1998wn} and Roget's Thesaurus \cite{Kipfer:2009book}, comprise a total of 1,023,082 synonymy/\textsc{attract} word pairs and 380,873 antonymy/\textsc{repel} pairs. 

Note that the sets of constraints cover only a fraction of the full distributional vocabulary, providing direct motivation for post-specialization methods which are able to specialize the full vocabulary. For instance, only 15.3\% of the \textsc{sgns-w2} vocabulary words are seen words present in the constraints.\footnote{The respective coverage for the 200K most frequent \textsc{glove-cc} and \textsc{fasttext} words is only 13.3\% and 14.6\%.}

The constraints are initially injected into the distributional vector space (see Figure~\ref{fig:illustration} again) using \textsc{attract-repel}, a state-of-the-art specialization model, for which we adopt the original suggested model setup \cite{Mrksic:2017tacl}.\footnote{\url{https://github.com/nmrksic/attract-repel}} Hyper-parameter values are set to: $\delta_{A}=0.6$, $\delta_{R}=0.0$, $\lambda_{P}=10^{-9}$. The models are trained for 5 epochs with Adagrad \cite{Duchi:11}, with batch sizes set to $k_{A}=k_{R}=50$, again as in the original work. 

\paragraph{AuxGAN Setup and Hyper-Parameters.}
Both the generator and the discriminator are feed-forward nets with $l=2$ hidden layers, each of size $h=2048$, and LeakyReLU as non-linear activation \cite{Maas:2014icml}. The dropout for the input and hidden layers of the generator is 0.2 and for the input layer of the discriminator 0.1. In evaluation, the noise is blanketed out in order to ensure a deterministic mapping \citep{isola2017image}. Moreover, we smooth the golden labels for prediction by a factor of 0.1 to make the model less vulnerable to adversarial examples \citep{szegedy2016rethinking}.

We train our model with SGD for 10 epochs of 1 million iterations each, feeding mini-batches of size 32. For each pair in a batch we generate 25 negative examples; $s=5$ (see \S\ref{ssec:gan}). As a way to normalize the mini-batches \citep{salimans2016improved}, these are constructed to contain exclusively either original or specialized vectors. At each epoch, the initial learning rate of 0.1 is decayed by a factor of 0.98, or 0.5 if the score on the validation set (computed as the average cosine similarity between the predicted and \textsc{ar}-specialized embeddings)\footnote{The score is computed as the average cosine similarity between the original and specialized embeddings. 
} 
has not increased. The hyper-parameters $k$ and $\delta_{MM}$ are tuned via grid search on the validation set. 


\paragraph{Zero-Shot Specialization Setup.}
The GAN discriminator for learning a shared cross-lingual vector space (see \S\ref{ssec:biling}) has hyper-parameters identical to the AuxGAN. The generator instead is a linear layer initialized as an identity matrix and enforced to lie on the manifold of orthogonal matrices during training \citep{cisse2017parseval}. No dropout is used. The unsupervised validation metric for early stopping is the cosine distance between dictionary pairs extracted with the CSLS similarity metric.


%% file: 05_results.tex
\subsection{Word Similarity}
\label{ss:ws}
\paragraph{Evaluation Setup.}
We first evaluate adversarial post-specialization intrinsically, using two standard word similarity benchmarks for English: SimLex-999 \cite{Hill:2015cl} and SimVerb-3500 \cite{Gerz:2016emnlp}, a dataset containing human similarity ratings for 3,500 verb pairs.\footnote{Unlike WordSim-353 \cite{Finkelstein:2002tois} or MEN \cite{Bruni:2014jair}, SimLex and SimVerb provide explicit guidelines to discern between true semantic similarity and (more broad) conceptual relatedness, so that related but non-similar words (e.g. \textit{tiger} and \textit{jungle}) have a low rating.} The evaluation measure is Spearman's $\rho$ rank correlation between gold and predicted word pair similarity scores.

We evaluate word vectors in two settings, similar to \newcite{Vulic:2018naaclpost}. \textbf{a)} In the synthetic \textsc{disjoint} setting, we discard all linguistic constraints that contain any of the words found in SimLex or SimVerb. This means that all test words from SimLex and SimVerb are effectively unseen words, and through this setting we are able to \textit{in vitro} evaluate the model's ability to generalize the specialization function to unseen words. \textbf{b)} In the \textsc{full} setting we leverage all constraints. This is a standard ``real-life'' scenario where some test words do occur in the constraints, while the mapping is learned for the remaining words. We use the \textsc{full} setting in all subsequent downstream applications (\S\ref{ss:downstream}).

\setlength{\tabcolsep}{5.5pt}
\begin{table*}[!t]
\centering
\def\arraystretch{0.97}
\vspace{-0.0em}
{\footnotesize
\begin{tabularx}{\linewidth}{l cc cc cc cc cc cc}
\toprule
{} & \multicolumn{6}{c}{Setting: \textsc{disjoint}} & \multicolumn{6}{c}{Setting: \textsc{full}} \\
\cmidrule(lr){2-7} \cmidrule(lr){8-13}
{} & \multicolumn{2}{c}{\textsc{glove-cc}} & \multicolumn{2}{c}{\textsc{fasttext}} & \multicolumn{2}{c}{\textsc{sgns-w2}} & \multicolumn{2}{c}{\textsc{glove-cc}} & \multicolumn{2}{c}{\textsc{fasttext}} & \multicolumn{2}{c}{\textsc{sgns-w2}} \\
\cmidrule(lr){2-7} \cmidrule(lr){8-13}
{} & {SL} & {SV} & {SL} & {SV} & {SL} & {SV} & {SL} & {SV} & {SL} & {SV} & {SL} & {SV} \\
\cmidrule(lr){2-3} \cmidrule(lr){4-5} \cmidrule(lr){6-7} \cmidrule(lr){8-9} \cmidrule(lr){10-11} \cmidrule(lr){12-13}
{\textbf{Distributional} ($\mathbf{X}$)} & {.407} & {.280} & {.383} & {.247} & {.414} & {.272} & {.407} & {.280} & {.383} & {.247} & {.414} & {.272} \\
\textbf{Specialized:} \textsc{Attract-Repel} & {.407} & {.280} & {.383} & {.247} & {.414} & {.272} & {.781} & {.761} & {.764} & {.744} & {.778} & {.761} \\
{\textbf{Post-Specialized}: \textsc{post-dffn}} & {.645} & {.531} & {.503} & {.340} & {.553} & {.430} & {.785} & {.764} & {.768} & {.745} & {.781} & {.763} \\
{\textbf{Post-Specialized}: \textsc{auxgan}} & \textbf{.652}	& \textbf{.552} & {\textbf{.513}} & {\textbf{.394}} & {\textbf{.581}} & {\textbf{.434}} & .789 & .764 & .766 &.741 & .782 & .762 \\
	
\bottomrule
\end{tabularx}}

\caption{Spearman's $\rho$ correlation scores for three standard English distributional vectors spaces on English SimLex-999 (SL) and SimVerb-3500 (SV). \textsc{post-dffn} \cite{Vulic:2018naaclpost} uses a deep non-linear feed-forward network to learn the mapping function $f$. \textsc{auxgan} is our adversarial model (see \S\ref{ssec:gan}).}
\label{tab:ws}
\end{table*}

We compare our model to \textsc{Attract-Repel (ar)}, which specializes only the vectors of words occurring in the constraints. We also provide comparisons to a post-specialization model of \newcite{Vulic:2018naaclpost} which specializes the full vocabulary, but substitutes the AuxGAN architecture from \S\ref{ssec:gan} with a deep $5$-layer feed-forward neural net also based on the max-margin loss (see Eq.~\eqref{eq:mm}) to learn the mapping function (\textsc{post-dffn}).


\paragraph{Results and Analysis.} 
The results are summarized in Table~\ref{tab:ws}. The scores suggest that the proposed adversarial post-specialization model is universally useful and robust: we observe gains over input distributional word vectors for all three vector collections. The results in the \textsc{disjoint} setting illustrate the core limitation of the initial specialization/post-processing models and indicate the extent of improvement achieved when generalizing the specialization function to unseen words through adversarial post-specialization. Moreover, the scores suggest that the more sophisticated adversarial post-specialization method (\textsc{auxgan}) outperforms \textsc{post-dffn} across a large number of experimental runs, verifying its effectiveness. 

We observe only modest and inconsistent gains over \textsc{attract-repel} and \textsc{post-dffn} in the \textsc{full} setting. However, the explanation of this finding is straightforward: 99.2\% of SimLex words and 99.9\% of SimVerb words are present in the external constraints, making this an unrealistic evaluation scenario. The usefulness of the initial \textsc{attract-repel} specialization is less pronounced in real-life downstream applications in which such high coverage cannot be guaranteed, as shown in \S\ref{ss:downstream}.

\subsection{Downstream Tasks}
\label{ss:downstream}

We next evaluate the embedding spaces specialized with the AuxGAN method in two tasks in which discerning semantic similarity from semantic relatedness is crucial: lexical text simplification (LS) and dialog state tracking (DST). 

\subsubsection{Lexical Text Simplification}

The goal of lexical simplification is to replace complex words (typically words that are used less often in language and are therefore less familiar to readers) with their simpler synonyms, without infringing the grammaticality and changing the meaning of the text. Replacing complex words with related words instead of true synonyms affects the original meaning (e.g., \textit{Ferrari pilot Vettel} vs \textit{Ferrari airplane Vettel}) and often yields ungrammatical text (e.g., \textit{they drink all pizzas}). 

\paragraph{LS Using Word Vectors.} We use Light-LS, a publicly available LS tool based on word embeddings \cite{Glavas:2015acl}. Light-LS generates and then ranks substitution candidates based on similarity in the input word vector space. The quality of the space thus directly affects LS performance: by plugging any word vector space into Light-LS, we extrinsically evaluate that embedding space for LS. Furthermore, the better the embedding space captures true semantic similarity, the better the substitutions made by Light-LS.      

\paragraph{Evaluation Setup.} We use the standard LS dataset of \newcite{horn2014learning}. It contains 500 sentences with indicated complex words (one word per sentence) that have to be substituted with simpler synonyms. For each word, simplifications were crowdsourced from 50 human annotators. Following prior work \cite{horn2014learning,Glavas:2015acl}, we evaluate the performance of Light-LS using the metric that quantifies both the quality and the frequency of word replacements: \textit{Accurracy (Acc)} metric is the number of correct simplifications made divided by the total number of complex words. 

\setlength{\tabcolsep}{3.8pt}
\begin{table}[t]
\centering
\def\arraystretch{0.97}
\vspace{-0.0em}
{\footnotesize
\begin{tabularx}{\linewidth}{l ccc}
\toprule
{} & \multicolumn{1}{c}{\textsc{glove-cc}} & \multicolumn{1}{c}{\textsc{fasttext}} & \multicolumn{1}{c}{\textsc{sgns-w2}} \\
\cmidrule(lr){2-2} \cmidrule(lr){3-3} \cmidrule(lr){4-4}
{Vector space} & {Acc} & {Acc} & {Acc} \\ \midrule
\textbf{Distributional} & {.660} & {.578} & {.560}  \\ 
\textbf{Specialized}: \textsc{ar} & {.676}  & {.698}  & {.644}  \\
\textbf{Post-Specialized}:  & {} & {} & {} \\
{\textsc{post-dffn}} & {\bf .723} & {.723} & {.709} \\
{\textsc{auxgan}} & {.717} & {\bf .739} & {\bf .721} \\
\bottomrule
\end{tabularx}}
\vspace{-0.5mm}
\caption{Lexical simplification results for three (post-specialized) distributional spaces.}
\label{tbl:simplification}
\end{table}

\paragraph{Results and Analysis.} 
Scores for all three pre-trained vector spaces are shown in Table \ref{tbl:simplification}. Similar to the word similarity task, embedding spaces produced with post-specialization models outperform the vectors produced with \textsc{ar} and original distributional vectors. The gains are now more pronounced in the real-life \textsc{full} setup, as only 59.6 \% of all indicated complex words and substitution candidates from the LS dataset are covered in the external constraints. Adversarial post-specialization (\textsc{auxgan}) has a slight edge over the post-specialization with a simple feed-forward network (\textsc{post-dffn}) for \textsc{fasttext} and \textsc{sgns-w2} embeddings, but not for \textsc{glove-cc} vectors. In general, the fact that both post-specialization methods outperform \textsc{attract-repel} by a wide margin shows the importance of specializing the full word vector space for downstream NLP applications.       

\subsubsection{Dialog State Tracking}
Finally, we evaluate the importance of full-vocabulary (adversarial) post-specialization in another language understanding task: dialog state tracking (DST) \cite{Henderson:14a,Williams:16}, which is a standard task to measure the impact of specialization in prior work \cite{Mrksic:2017tacl}. A DST model is typically the first component of a dialog system pipeline \cite{young:10}, tasked with capturing user's goals and updating the dialog belief state at each dialog turn. Distinguishing similarity from relatedness is crucial for DST (e.g., a dialog system should not recommend an \textit{``expensive restaurant in the west''} when asked for an \textit{``affordable pub in the north''}).

\paragraph{Evaluation Setup.} To evaluate the effects of specialized word vectors on DST, following prior work we utilize the Neural Belief Tracker (NBT), a statistical DST model that makes inferences purely based on pre-trained word vectors \cite{Mrksic:2017acl}.\footnote{\url{https://github.com/nmrksic/neural-belief-tracker}; For full model details, we refer the reader to the original paper.} Again, as in prior work the DST evaluation is based on the Wizard-of-Oz (WOZ) v2.0 dataset \cite{Wen:17,Mrksic:2017acl}, comprising 1,200 dialogues split into training (600 dialogues), development (200), and test data (400). We report the standard DST metric: \textit{joint goal accuracy (JGA)}, the proportion of dialog turns where all the user's search goal constraints were correctly identified, computed as average over 5 NBT runs.
\setlength{\tabcolsep}{20pt}
\begin{table}[t]
\centering
\def\arraystretch{0.98}
\vspace{-0.0em}
{\footnotesize
\begin{tabularx}{\linewidth}{l r}
\toprule
\textsc{glove-cc} word vectors & JGA \\ \midrule 
\textbf{Distributional} & {.797} \\ 
\textbf{Specialized}: \textsc{attract-repel} & {.817} \\
\textbf{Post-Specialized}: \textsc{post-dffn} & {.829}  \\
\textbf{Post-Specialized}: \textsc{auxgan} & \textbf{.836}  \\
\bottomrule
\end{tabularx}}
\vspace{-0.5mm}
\caption{English DST performance (joint goal accuracy). \textsc{GloVe-CC} word vectors.}
\label{tab:dst}
\vspace{-1.5mm}
\end{table}

\paragraph{Results and Analysis.} We show English DST performance in the \textsc{full} setting in Table~\ref{tab:dst}. Only NBT performance with \textsc{glove-cc} vectors is reported for brevity, as similar performance gains are observed with the other two pre-trained vector collections. The results confirm our findings established in the other two tasks: \textbf{a)} initial \textsc{ar} specialization of distributional vectors is useful, but \textbf{b)} it is crucial to specialize the full vocabulary for improved performance (e.g., 57\% of all WOZ words are present in the constraints), and \textbf{c)} the more sophisticated \textsc{auxgan} model yields additional gains.


\setlength{\tabcolsep}{3.8pt}
\begin{table}[t]
\centering
\def\arraystretch{0.98}
\vspace{-0.0em}
{\footnotesize
\begin{tabularx}{\linewidth}{l XXX XXX}
\toprule
{} & \multicolumn{2}{c}{Similarity ($\rho$)} & \multicolumn{2}{c}{LS (Acc)} & \multicolumn{2}{c}{DST (JGA)}\\
\cmidrule(lr){2-3} \cmidrule(lr){4-5} \cmidrule(lr){6-7} 
{Vector space} & {IT} & {DE} & {IT} & {DE} & {IT} & {DE} \\
\cmidrule(lr){2-3} \cmidrule(lr){4-5} \cmidrule(lr){6-7}
{\textbf{Distrib.}} & {.297} & {.417} & {.308} & {-} & {.681} & {.621} \\
{\textsc{auxgan}}& {\bf .431} & {\bf .525} & {\bf .392} & {-} & {\bf .714} & {\bf .651} \\
\bottomrule
\end{tabularx}}
\caption{Results of zero-shot specialization applied to IT and DE \textsc{fasttext} distributional vectors.}
\label{tab:zeroshot}
\vspace{-1.5mm}
\end{table}

\subsection{Cross-Lingual Zero-Shot Specialization}
\label{ss:zero-res}
\paragraph{Evaluation Setup.} Large collections of linguistic constraints do not exist for many languages. Therefore, we test if the specialization knowledge from a resoure-rich language (i.e., English) can be transferred to resource-lean target languages (see \S\ref{ssec:biling}). We simulate resource-lean scenarios using two target languages: Italian (IT) and German (DE).\footnote{Note that the two languages are not resource-poor, but we treat them as such in our experiments. This choice of languages was determined by the availability of high-quality evaluation data to measure the effects of zero-shot specialization.} We evaluate zero-specialized IT and DE \textsc{fasttext} vectors, using English \textsc{fasttext} vectors as the source, on the same three tasks as before. We report the same evaluation measures, using the following evaluation data: 1) IT and DE SimLex-999 datasets \cite{Leviant:2015arxiv} for word similarity; 2) IT lexical simplification data  (SIMPITIKI) \cite{Tonelli:2016simpitiki}; 3) IT and DE WOZ data \cite{Mrksic:2017tacl} for DST. 


\paragraph{Results and Analysis.} 
The results are summarized in Table~\ref{tab:zeroshot}. The gains over the original distributional vectors are substantial across all three tasks and for both languages. This finding indicates that the semantic content of distributional vectors can be enriched even for languages without any readily available lexical resources. 

The gap between performances of language transfer and the monolingual setting is explained by the noise introduced by the bilingual vector alignment and the different ways concepts are lexicalized across languages, as studied by semantic typology \citep{ponti2018modeling}.
Nonetheless, in the long run, these transfer results hold promise to support the specialization of vector spaces even for resource-lean languages, and their applications. 

%% file: 02_related.tex

\textbf{Vector Space Specialization.} 
Specialization methods embed external information into vector spaces. 
Some of them integrate external linguistic constraints into distributional training and \textit{jointly} optimize distributional and non-distributional objectives: they modify the prior or the regularization \cite{Yu:2014,Xu:2014,Bian:14,Kiela:2015emnlp}, or use a variant of the SGNS-style objective \cite{Liu:EtAl:15,Ono:2015naacl,Osborne:16}. 

Other models inject external knowledge from available lexical resources (e.g., WordNet, PPDB) into pre-trained word vectors as a \textit{post-processing step} \cite{Faruqui:2015naacl,Rothe:2015acl,Wieting:2015tacl,Nguyen:2016acl,Mrksic:2016naacl,Cotterell:2016acl,Mrksic:2017tacl}. They offer a portable, flexible, and light-weight approach to incorporating external knowledge into \textit{arbitrary} vector spaces, outperforming less versatile joint models and yielding state-of-the-art results on language understanding tasks \cite{Mrksic:2016naacl,Kim:2016slt,Vulic:2017acl}. By design, these methods fine-tune only vectors of words {seen} in external resources.

\newcite{Vulic:2018naaclpost} suggest that specializing the full vocabulary is beneficial for downstream applications. Comparing to their work, we show that a more sophisticated adversarial post-specialization can yield further gains across different tasks and boost full-vocabulary specialization in resource-lean settings through cross-lingual transfer.

\vspace{1.4mm}
\noindent \textbf{Generative Adversarial Networks.} GANs were originally devised to generate images from input noise variables \citep{goodfellow2014generative}. The generation process is typically conditioned on discrete labels or data from other modalities, such as text \citep{mirza2014conditional}. Otherwise, the condition can take the form of real data in input rather than (or in addition to) noise: in this case, the generator parameters are better conceived as a mapping function. For instance, it can bridge between pixel-to-pixel \citep{isola2017image} or character-to-pixel \citep{reed2016generative} transformations. 

The GAN objective can be mixed with more traditional loss functions: in these cases, apart from trying to fool the discriminator, the generator also minimizes the distance between input and target data \cite{pathak2016context,li2016precomputed,ledig2017photo}. The distance can be formulated as the mean squared error between the input and the target \citep{pathak2016context}, their feature maps \citep{li2016precomputed}, both \citep{zhu2016generative}, or a loss calculated on feature maps of a deep convolutional network \citep{ledig2017photo}.

In the textual domain, adversarial models have been proven to support domain adaptation \citep{ganin2016domain} and language transfer \citep{chen2016adversarial} by learning domain/language-invariant latent features. Adversarial training also powers unsupervised mapping between monolingual vector spaces to learn cross-lingual word embeddings \cite{zhang2017adversarial,Conneau:2018iclr}. In this work, we show how to apply adversarial techniques to the problem of vector specialization, which has a substantial impact on language understanding tasks.


%% file: 06_conclusions.tex
We have presented adversarial post-specialization, a novel model supported by adversarial training which specializes word vectors for the full vocabulary of the input distributional vector space, including words unseen in external lexical resources. We have also introduced a method for zero-shot specialization of word vectors in languages without any external resources. The benefits of adversarial post-specialization and its zero-shot transfer have been demonstrated across three tasks (word similarity, lexical text simplification, and dialog state tracking) and for three languages. 

In future work, we will explore more sophisticated adversarial models such as Cycle-GAN \cite{zhu2017unpaired}. Moreover, we will experiment with bootstrapping approaches to extract new lexical constraints from post-specialized embeddings. We also plan to extend the method to asymmetric relations (e.g., hypernymy) and to more target (resource-lean) languages. The code is available at \url{https://github.com/cambridgeltl/adversarial-postspec}.  
